\documentclass[runningheads]{llncs}
\usepackage{graphicx}

\usepackage{tikz}
\usepackage{comment}
\usepackage{amsmath,amssymb} 
\usepackage{color}
\usepackage{subcaption}
\usepackage{hyperref}

\usepackage[accsupp]{axessibility}  

\begin{document}
\pagestyle{headings}
\mainmatter
\def\ECCVSubNumber{364}  

\title{AIM 2022 Challenge on Instagram Filter Removal: Methods and Results}

\titlerunning{AIM 2022 Challenge on Instagram Filter Removal: Methods and Results}
%
\author{Furkan Kınlı \and Sami Menteş \and Barış Özcan \and Furkan Kıraç \and Radu Timofte \and 
        Yi Zuo \and Zitao Wang \and Xiaowen Zhang \and 
        Yu Zhu \and Chenghua Li \and Cong Leng \and Jian Cheng \and 
        Shuai Liu \and Chaoyu Feng \and Furui Bai \and Xiaotao Wang \and Lei Lei \and 
        Tianzhi Ma \and Zihan Gao \and Wenxin He \and 
        Woon-Ha Yeo \and Wang-Taek Oh \and Young-Il Kim \and Han-Cheol Ryu \and 
        Gang He \and Shaoyi Long \and 
        S. M. A. Sharif \and Rizwan Ali Naqvi \and Sungjun Kim 
        Guisik Kim \and Seohyeon Lee \and 
        Sabari Nathan \and Priya Kansal 
        }
%
\authorrunning{F. Kınlı, S. Menteş, B. Özcan, F. Kıraç, R, Timofte, et al.} 
%
\institute{
Özyeğin University, University of Würzburg, 
Xidian University, 
Chinese Academy of Sciences, MAICRO, AiRiA, 
Xiaomi Inc., 
Sahmyook University, 
FS Solution, Sejong University, 
Chung-Ang University, 
Couger Inc. 
}
\maketitle

\begin{abstract}
This paper introduces the methods and the results of AIM 2022 challenge on Instagram Filter Removal. Social media filters transform the images by consecutive non-linear operations, and the feature maps of the original content may be interpolated into a different domain. This reduces the overall performance of the recent deep learning strategies. The main goal of this challenge is to produce realistic and visually plausible images where the impact of the filters applied is mitigated while preserving the content. The proposed solutions are ranked in terms of the PSNR value with respect to the original images. There are two prior studies on this task as the baseline, and a total of 9 teams have competed in the final phase of the challenge. The comparison of qualitative results of the proposed solutions and the benchmark for the challenge are presented in this report.
\end{abstract}

\section{Introduction}

Understanding the content of social media images is a crucial task for industrial applications in different domains. The images shared on social media platforms generally contain visual filters applied where these filters lead to different consecutive transformations on top of the images or inject varied combinations of distractive factors like noise or blurring. Filtered images are the challenging subjects for the recent deep learning strategies (\textit{e.g.}, CNN-based architectures) since they are not robust to these transformations or distractive factors. These filters basically lead to interpolating the feature maps of the original content, and the final outputs for the original content and its filtered version are not the same due to this interpolation factor. Earlier studies addressing this issue present some prior solutions which focus on the filter classification \cite{bianco2017artistic,filterInvariant,chu2019photo,Wu_Wu_Singh_Davis_2020} or learning the inverse of a set of transformations applied \cite{artisticFilter,photoTransform}. Recent studies \cite{Kinli_2021_CVPR,Kinli_2022_CVPR} introduce the idea of recovering the original image from its filtered version as pre-processing in order to improve the overall performance of the models employed for visual understanding of the content. 

We can define that the main goal of Instagram filter removal is to produce realistic and pleasant images in which the impact of the filters applied is mitigated while preserving the original content within. Jointly with the AIM workshop, we propose an AIM challenge on Instagram Filter Removal: the task of recovering the set of Instagram-filtered images to their original versions with high fidelity. The main motivation in this challenge is to increase attention on this research topic, which may be helpful for the applications that primarily target social media images.

This challenge is a part of the AIM 2022 Challenges: Real-Time Image Super-Resolution~\cite{ignatov2022isr}, Real-Time Video Super-Resolution~\cite{ignatov2022vsr}, Single-Image Depth Estimation~\cite{ignatov2022depth}, Learned Smartphone ISP~\cite{ignatov2022isp}, Real-Time Rendering Realistic Bokeh~\cite{ignatov2022bokeh}, Compressed Input Super-Resolution~\cite{yang2022aim} and Reversed ISP~\cite{conde2022aim}. The results obtained in the other competitions and the description of the proposed solutions can be found in the corresponding challenge reports.

\section{Challenge}

In this challenge, the participants are expected to provide a solution that removes Instagram filters from the given images. Provided solutions are ranked in terms of the Peak Signal-to-Noise Ratio (PSNR) of their outputs with respect to the ground truth. Each entry was required to submit the model file along with a runnable code file and the outputs obtained by the model. 

\subsection{Challenge Data}  

We have used the IFFI dataset \cite{Kinli_2021_CVPR} in this challenge, which has 500 and 100 instances of original images and their filtered versions as the training and validation sets, respectively. IFFI dataset\footnote{\url{https://github.com/birdortyedi/instagram-filter-removal-pytorch}} are provided for the training and validation phases in this challenge. For the final evaluation phase, we have shared another 100 instances of images with 11 filters (\textit{i.e.} available for the annotation) for the private test set. We have followed the same procedure given in \cite{Kinli_2021_CVPR} during the annotation. The filters are picked among the ones in the training and validation sets, but available on Instagram at the time of annotation for the private test set, which are \textit{Amaro}, \textit{Clarendon}, \textit{Gingham}, \textit{He-Fe}, \textit{Hudson}, \textit{Lo-Fi}, \textit{Mayfair}, \textit{Nashville}, \textit{Perpetua}, \textit{Valencia}, and \textit{X-ProII}. All of the samples are originally collected in high-resolution (\textit{i.e.}, 1080px), then we have to downsample them to the low-resolution (\textit{i.e.}, 256px) in order to follow the previous benchmark given in \cite{Kinli_2021_CVPR}. To avoid anti-aliasing, gaussian blur is applied prior to the downsampling. High-resolution images are also provided to the participants in the case of using their pre-processing strategy. The final evaluation has been done on low-resolution images. The participants were free to use the extra data in their training process.

\subsection{Evaluation}

Considering Instagram Filter removal is an image-to-image translation problem where the images in a specific filter domain are translated into the output images in the original domain, we have measured the performance of the submissions by the common fidelity metrics in the literature (\textit{i.e.}, Peak Signal-to-Noise Ratio (PSNR) and the Structural Similarity (SSIM)). The participants have uploaded the outputs that they have obtained to CodaLab, and all evaluation process has been done on the CodaLab servers. We have shared the evaluation script on the challenge web page. In final evaluation phase, we ranked the submitted solutions by the average PSNR value on the private test set. Since it may be hard to differentiate the differences among the top solutions, we did not employ a user study in the evaluation process of this challenge.

\begin{table}[t]
\centering
\resizebox{\textwidth}{!}{%
\begin{tabular}{lllccccc}
\hline
\textbf{Team} & \textbf{Username} & \textbf{Framework} & \textbf{PSNR} & \textbf{SSIM} & \textbf{Runtime (s)} & \textbf{CPU/GPU} & \textbf{Extra Data} \\ \hline
Fivewin       & zuoyi             &     PyTorch               & \textbf{34.70}         & \textbf{0.97}          & 0.91                 & GPU              & No                  \\
CASIA LCVG    & zhuqingweiyu      &     PyTorch               & 34.48         & 0.96          & 0.43                 & GPU              & No                  \\
MiAlgo        & mialgo\_ls        &    PyTorch                & 34.47         & \textbf{0.97}          & 0.40                 & GPU              & No                  \\
Strawberry    & penwaterman       &    PyTorch                & 34.39         & 0.96          & 10.47                & GPU              & No                  \\
SYU-HnVLab    & Una               &  PyTorch           & 33.41         & 0.95          & 0.04                 & GPU              & No                  \\
XDER          & clearlon          &    PyTorch                & 32.19         & 0.95          & 0.05                 & GPU              & No                  \\
CVRG          & CVRG              &      PyTorch              & 31.78         & 0.95          & 0.06                 & GPU              & No                  \\
CVML          & sheee7            &      PyTorch              & 30.93         & 0.94          & \textbf{0.02}                 & GPU              & No                  \\
Couger AI     & SabariNathan      &     TensorFlow               & 30.83         & 0.94          & 10.43                & CPU              & Yes                 \\
IFRNet \cite{Kinli_2021_CVPR}       & Baseline 1        & PyTorch            & 30.46         & -             & 0.60                 & GPU              & No                  \\
CIFR \cite{Kinli_2022_CVPR}         & Baseline 2        & PyTorch            & 30.02         & -             & 0.62                 & GPU              & No                  \\ \hline
\end{tabular}%
}
\caption{The benchmark for the private test of IFFI dataset on Instagram Filter Removal challenge.}
\label{tab:benchmark}
\end{table}

\subsection{Submissions}

The participants were allowed to submit the unfiltered versions of the given images for 100 different instances in PNG format for validation and final evaluation phases. Each team could submit their outputs without any limits in the validation phase, while it is only allowed at most 3 times for the final evaluation phase. As mentioned before, we have created a private test set for the final evaluation phase and shared it with the participants without the original versions. After the submissions are completed, we have validated the leaderboard by double-checking the scores by the outputs sent in the submission files.

\section{Results}

This section introduces the benchmark results of Instagram Filter Removal challenge. It also presents the qualitative results of all proposed solutions. Lastly, the details of all solutions proposed for this challenge are described.

\captionsetup[subfigure]{font=small, labelformat=empty}
\begin{figure}[t]
        \centering
        \begin{subfigure}[b]{0.099\textwidth}
                \includegraphics[width=\textwidth]{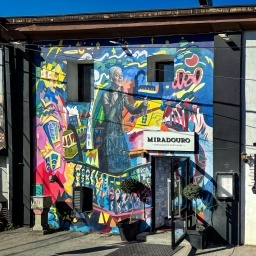}
                \includegraphics[width=\textwidth]{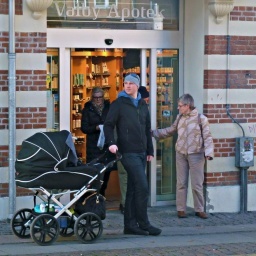}
                \includegraphics[width=\textwidth]{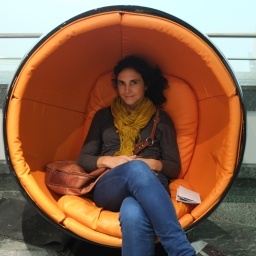}
                \includegraphics[width=\textwidth]{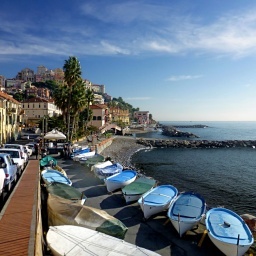}
                \includegraphics[width=\textwidth]{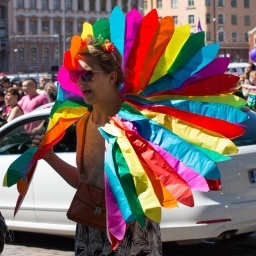}
                \includegraphics[width=\textwidth]{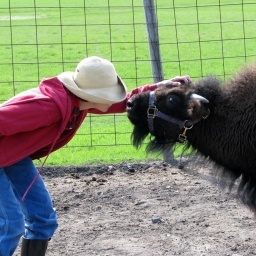}
                \includegraphics[width=\textwidth]{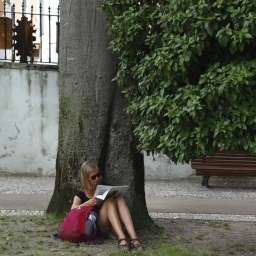}
                \includegraphics[width=\textwidth]{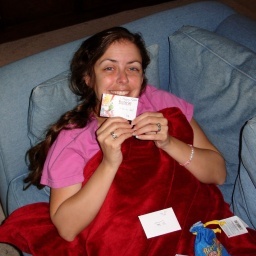}
                \label{fig:gt}
                \caption{GT}
        \end{subfigure}  
        \begin{subfigure}[b]{0.099\textwidth}
                \includegraphics[width=\textwidth]{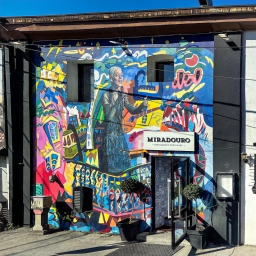}
                \includegraphics[width=\textwidth]{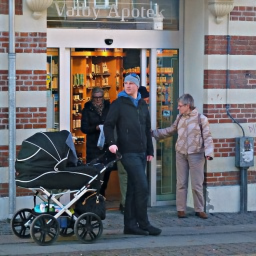}
                \includegraphics[width=\textwidth]{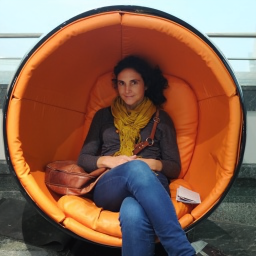}
                \includegraphics[width=\textwidth]{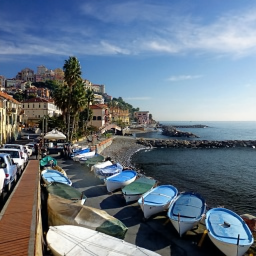}
                \includegraphics[width=\textwidth]{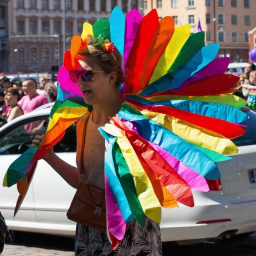}
                \includegraphics[width=\textwidth]{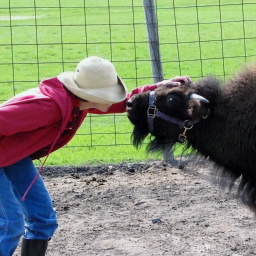}
                \includegraphics[width=\textwidth]{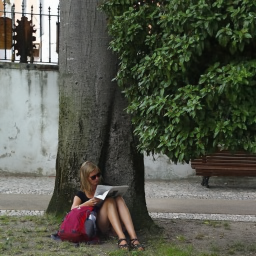}
                \includegraphics[width=\textwidth]{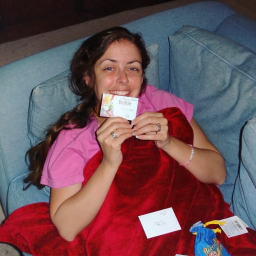}
                \label{fig:firewin}
                \caption{Fivewin}
        \end{subfigure}       
        \begin{subfigure}[b]{0.099\textwidth}
               \includegraphics[width=\textwidth]{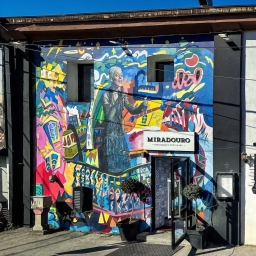}
                \includegraphics[width=\textwidth]{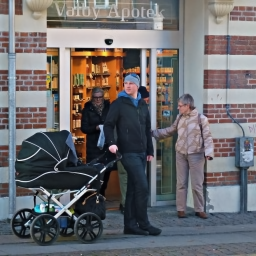}
                \includegraphics[width=\textwidth]{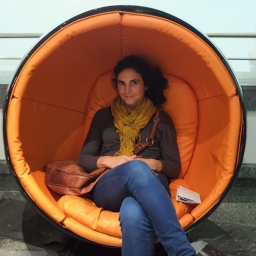}
                \includegraphics[width=\textwidth]{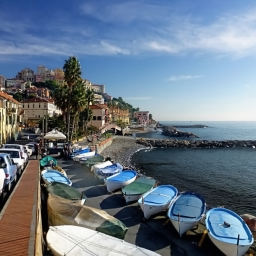}
                \includegraphics[width=\textwidth]{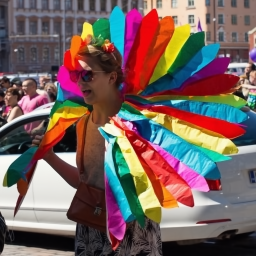}
                \includegraphics[width=\textwidth]{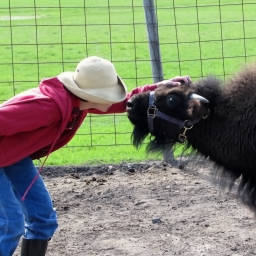}
                \includegraphics[width=\textwidth]{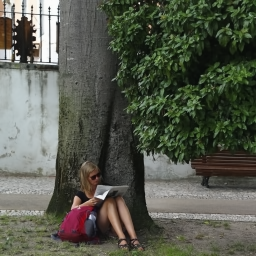}
                \includegraphics[width=\textwidth]{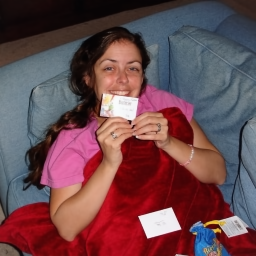}
                \label{fig:casia}
                \caption{CASIA}
        \end{subfigure}
        \begin{subfigure}[b]{0.099\textwidth}
                \includegraphics[width=\textwidth]{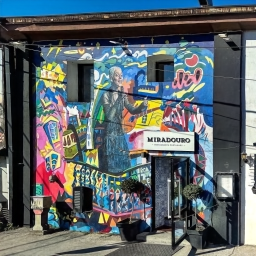}
                \includegraphics[width=\textwidth]{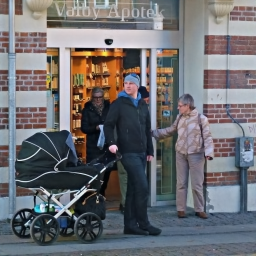}
                \includegraphics[width=\textwidth]{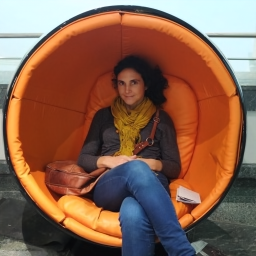}
                \includegraphics[width=\textwidth]{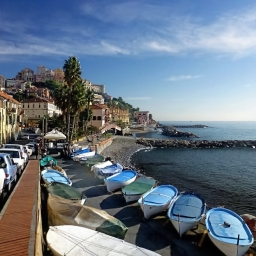}
                \includegraphics[width=\textwidth]{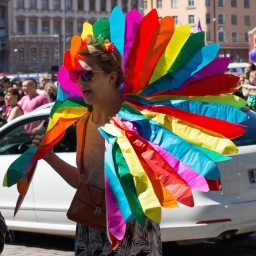}
                \includegraphics[width=\textwidth]{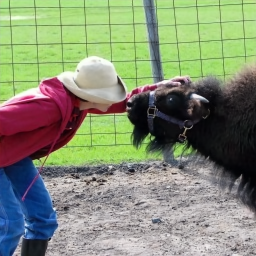}
                \includegraphics[width=\textwidth]{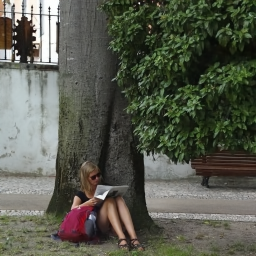}
                \includegraphics[width=\textwidth]{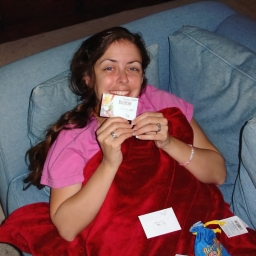}
                \label{fig:mi}
                \caption{MiAlgo}
        \end{subfigure}
        \begin{subfigure}[b]{0.099\textwidth}
                \includegraphics[width=\textwidth]{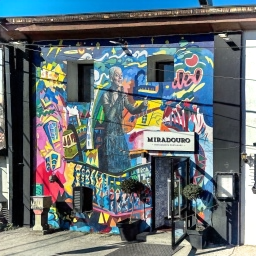}
                \includegraphics[width=\textwidth]{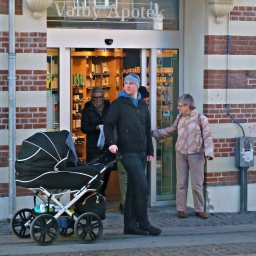}
                \includegraphics[width=\textwidth]{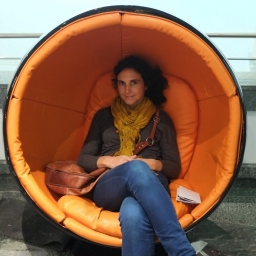}
                \includegraphics[width=\textwidth]{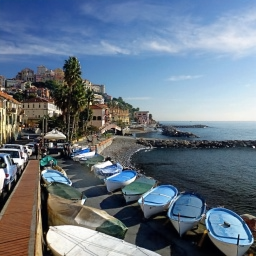}
                \includegraphics[width=\textwidth]{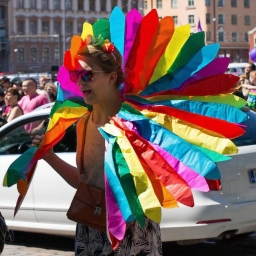}
                \includegraphics[width=\textwidth]{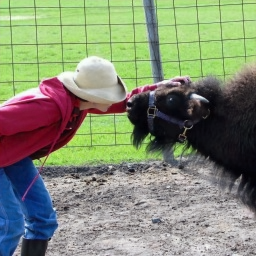}
                \includegraphics[width=\textwidth]{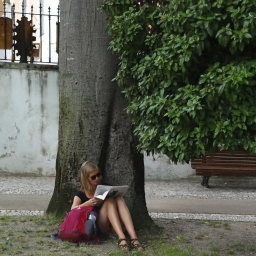}
                \includegraphics[width=\textwidth]{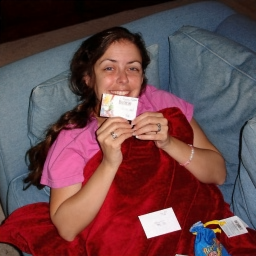}
                \label{fig:syu}
                \caption{SYU}
        \end{subfigure}
        \begin{subfigure}[b]{0.099\textwidth}
                \includegraphics[width=\textwidth]{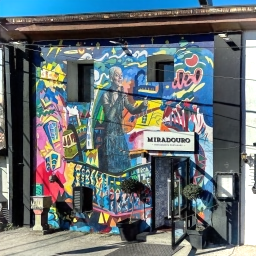}
                \includegraphics[width=\textwidth]{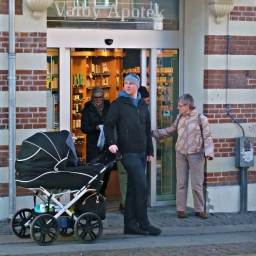}
                \includegraphics[width=\textwidth]{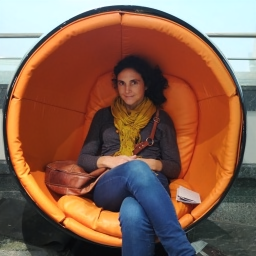}
                \includegraphics[width=\textwidth]{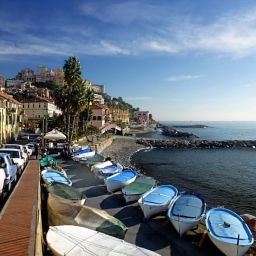}
                \includegraphics[width=\textwidth]{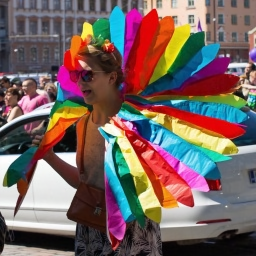}
                \includegraphics[width=\textwidth]{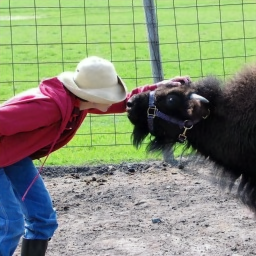}
                \includegraphics[width=\textwidth]{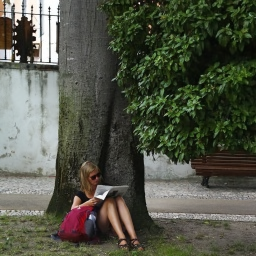}
                \includegraphics[width=\textwidth]{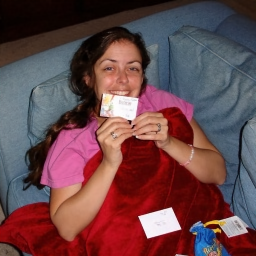}
                \label{fig:xder}
                \caption{XDER}
        \end{subfigure}
        \begin{subfigure}[b]{0.099\textwidth}
                \includegraphics[width=\textwidth]{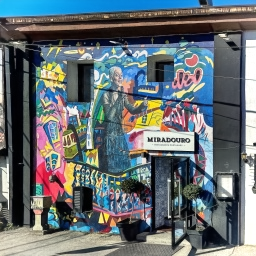}
                \includegraphics[width=\textwidth]{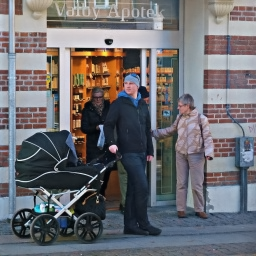}
                \includegraphics[width=\textwidth]{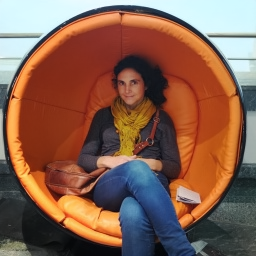}
                \includegraphics[width=\textwidth]{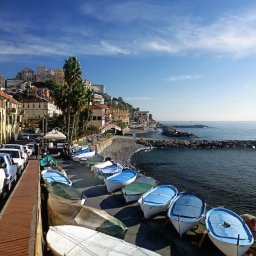}
                \includegraphics[width=\textwidth]{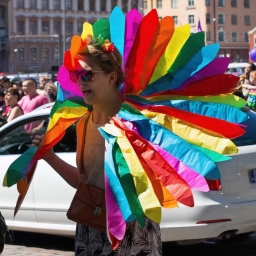}
                \includegraphics[width=\textwidth]{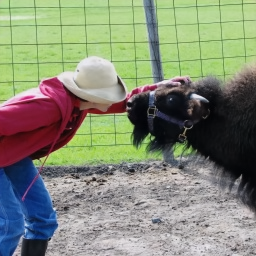}
                \includegraphics[width=\textwidth]{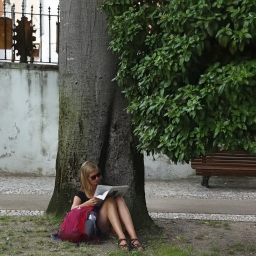}
                \includegraphics[width=\textwidth]{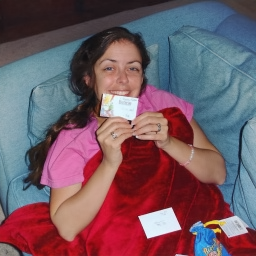}
                \label{fig:cvrg}
                \caption{CVRG}
        \end{subfigure}
        \begin{subfigure}[b]{0.099\textwidth}
                \includegraphics[width=\textwidth]{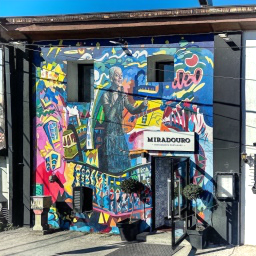}
                \includegraphics[width=\textwidth]{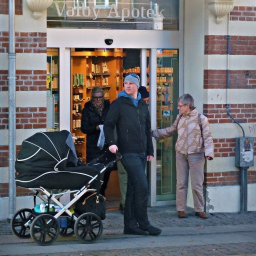}
                \includegraphics[width=\textwidth]{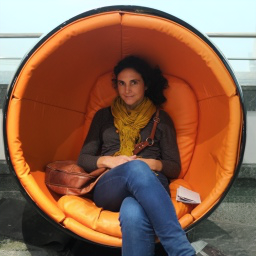}
                \includegraphics[width=\textwidth]{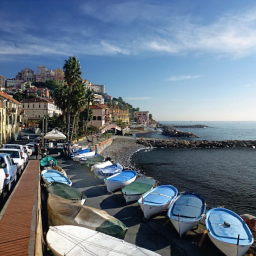}
                \includegraphics[width=\textwidth]{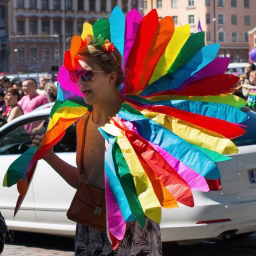}
                \includegraphics[width=\textwidth]{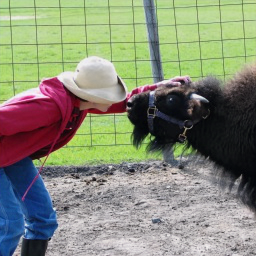}
                \includegraphics[width=\textwidth]{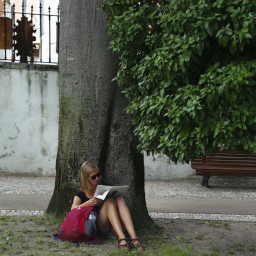}
                \includegraphics[width=\textwidth]{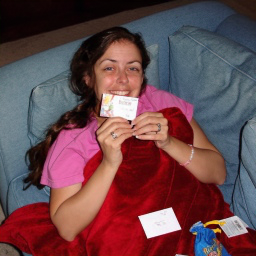}
                \label{fig:cvml}
                \caption{CVML}
        \end{subfigure}
        \begin{subfigure}[b]{0.099\textwidth}
                \includegraphics[width=\textwidth]{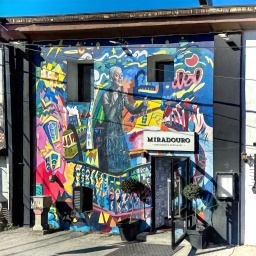}
                \includegraphics[width=\textwidth]{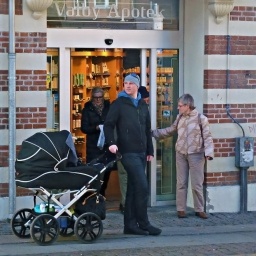}
                \includegraphics[width=\textwidth]{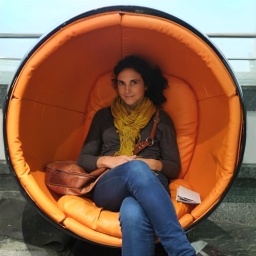}
                \includegraphics[width=\textwidth]{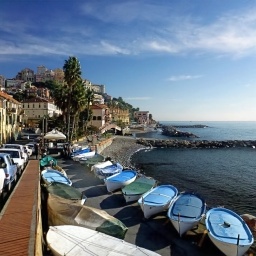}
                \includegraphics[width=\textwidth]{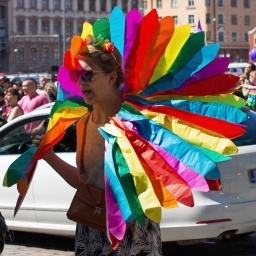}
                \includegraphics[width=\textwidth]{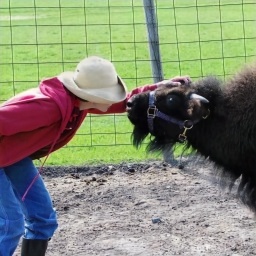}
                \includegraphics[width=\textwidth]{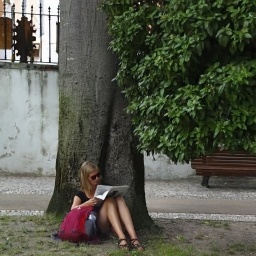}
                \includegraphics[width=\textwidth]{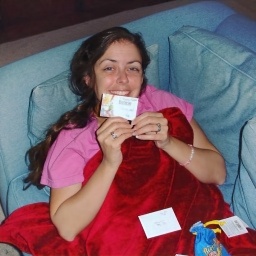}
                \label{fig:couger}
                \caption{Couger}
        \end{subfigure}
         \caption{Qualitative results on the private test set, submitted by the participants. From top to bottom, filter types: \textit{X-ProII}, \textit{He-Fe}, \textit{Amaro}, \textit{Perpetua}, \textit{Mayfair}, \textit{Gingham}, \textit{Lo-Fi}, \textit{Clarendon}; image IDs: 38, 14, 47, 88, 63, 11, 23, 55.}
         \label{fig:qual} 
\end{figure}

\subsection{Overall results}

From 114 registered participants, 9 teams entered the final phase and submitted the valid results, codes, executables, and factsheets. Table \ref{tab:benchmark} summarizes the final challenge results and presents PSNR and SSIM scores of all submitted solutions on the private test set. Figure \ref{fig:qual} demonstrates some examples among the qualitative results of all submitted solutions. Top solutions follow different deep learning strategies. The first team in the rankings has employed Transformer-based \cite{wang2022uformer} with a test-time augmentation mechanism. The team in the second rank has proposed a simple encoder-decoder architecture and a specialized defiltering module. They mainly focus on the training settings, instead of the network structure. Next, the MiAlgo team has used a two-stage learning approach where the first stage is responsible for processing the low-resolution input images and the second stage refines the output by using residual groups and different attention modules. Multi-scale color attention, style-based Transformer architecture, gated mechanism, two-branch approach, and a UNet-like architecture represent the other solutions followed by the rest of the participants.

\subsection{Solutions}

\subsubsection{Fivewin}
\begin{figure*}[t]
    \centering
    \begin{subfigure}[b]{0.49\textwidth}
        \centering
        \includegraphics[width=\textwidth]{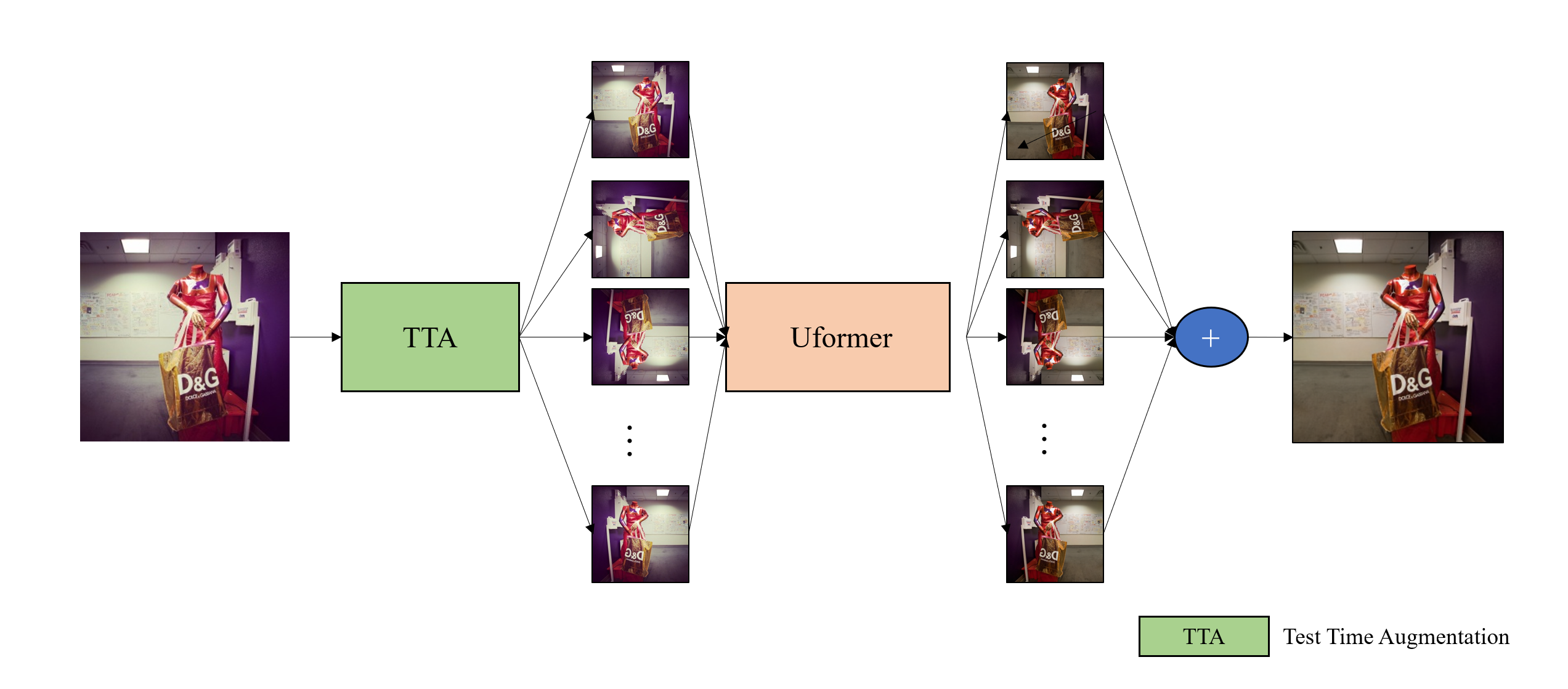}
        \caption{(a)}
    \end{subfigure}
    \begin{subfigure}[b]{0.49\textwidth}
        \centering
        \includegraphics[width=\textwidth]{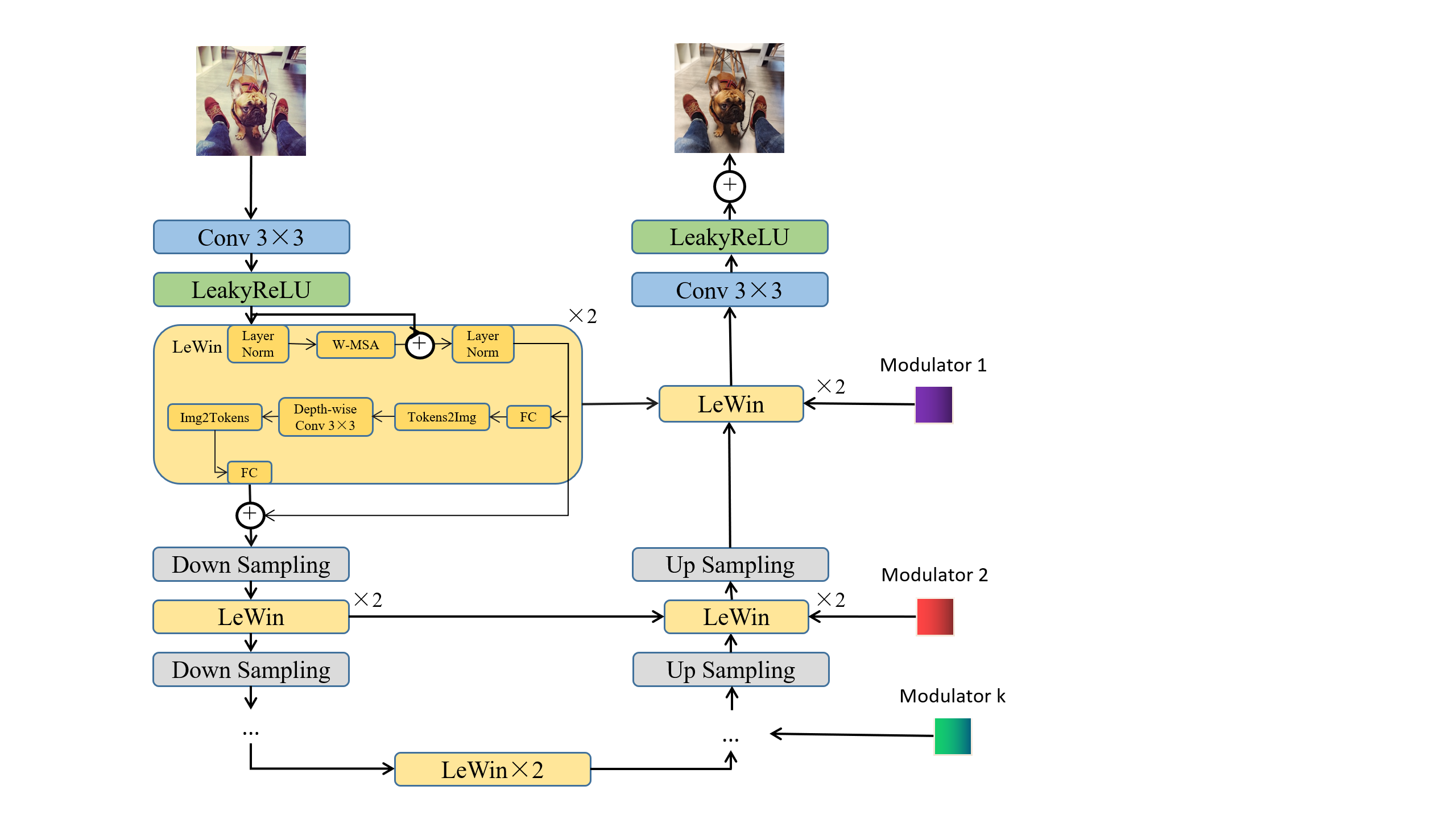}
        \caption{(b)}
    \end{subfigure}
    \caption{The proposed solution by Fivewin Team. \textbf{(a)} The main structure of UTFR is shown in this figure.. \textbf{(b)} The main structure of Uformer is shown in this figure.}
    \label{fig:Fivewin}
\end{figure*}

The Fivewin team proposed the UTFR method to restore the image after filtering (A U-shaped transformer for filter removal, abbreviated: UTFR). Figure \ref{fig:Fivewin} illustrates the main structure of the UTFR. We consider the filter as alternative image noise, so removing the filter is equivalent to denoising the image. We introduce the TTA mechanism (Test-Time Augmentation) based on the Uformer \cite{wang2022uformer} model to expand the amount of data and enhance the generalization of the model by data augmentation during the training of the model. Finally, the same data enhancement method is applied to the test images separately in the test and the changed images are fused to get the final results.

UTFR accepted an image size of $256\times256\times3$, trained for 250 epochs, with an initial learning rate of 0.0002, using the AdamW optimizer\cite{loshchilov2017decoupled} ($\beta_{1}=0.9, \beta_{2}=0.999$, eps$={1e}^{-8}$ , weight-decay$=0.02$). A cosine annealing method was used to perform the learning rate decay to ${1e}^{-6}$. The learning rate was gradually warmed up (\textit{i.e.,} increased) in the optimizer by warm-up, with the batch size set to 6. We used the Charbonnier loss function that can better handle outliers to approximate the $\mathcal{L}_1$ loss to improve the performance of the model, and finally used the PSNR as the evaluation metric for image restoration.

\subsubsection{CASIA LCVG}
The CASIA LCVG team proposed a simple encoder-decoder network for the Instagram Filter Removal task. As shown in Figure~\ref{DefilterNet}, DefilterNet consists of the basic Unit Defilter Block (DFB). Instead of focusing on designing complex network structures, this work tackles the problem from a different perspective, namely the training setting. They only use the normal end-to-end training method and try to enhance DFilterNet by investigating the training settings. Especially, they found that data augmentation plays an important role in this training method, especially Mixup~\cite{zhang2017mixup}, which effectively improves the DefilterNet by 2dB.

They train DefilterNet with AdamW optimizer ($\beta_{1}=0.9, \beta_{2}=0.9)$ and L1 loss for $300K$ iterations with the initial learning rate ${2e}^{-4}$ gradually reduced to ${1e}^{-6}$ with the cosine annealing. The training image size is  $256\times256$ and batch size is 24. For data augmentation, we use horizontal and vertical flips, especially involves Mixup~\cite{zhang2017mixup}.

\begin{figure}[t]
   \begin{center}
     \includegraphics[scale=0.35]{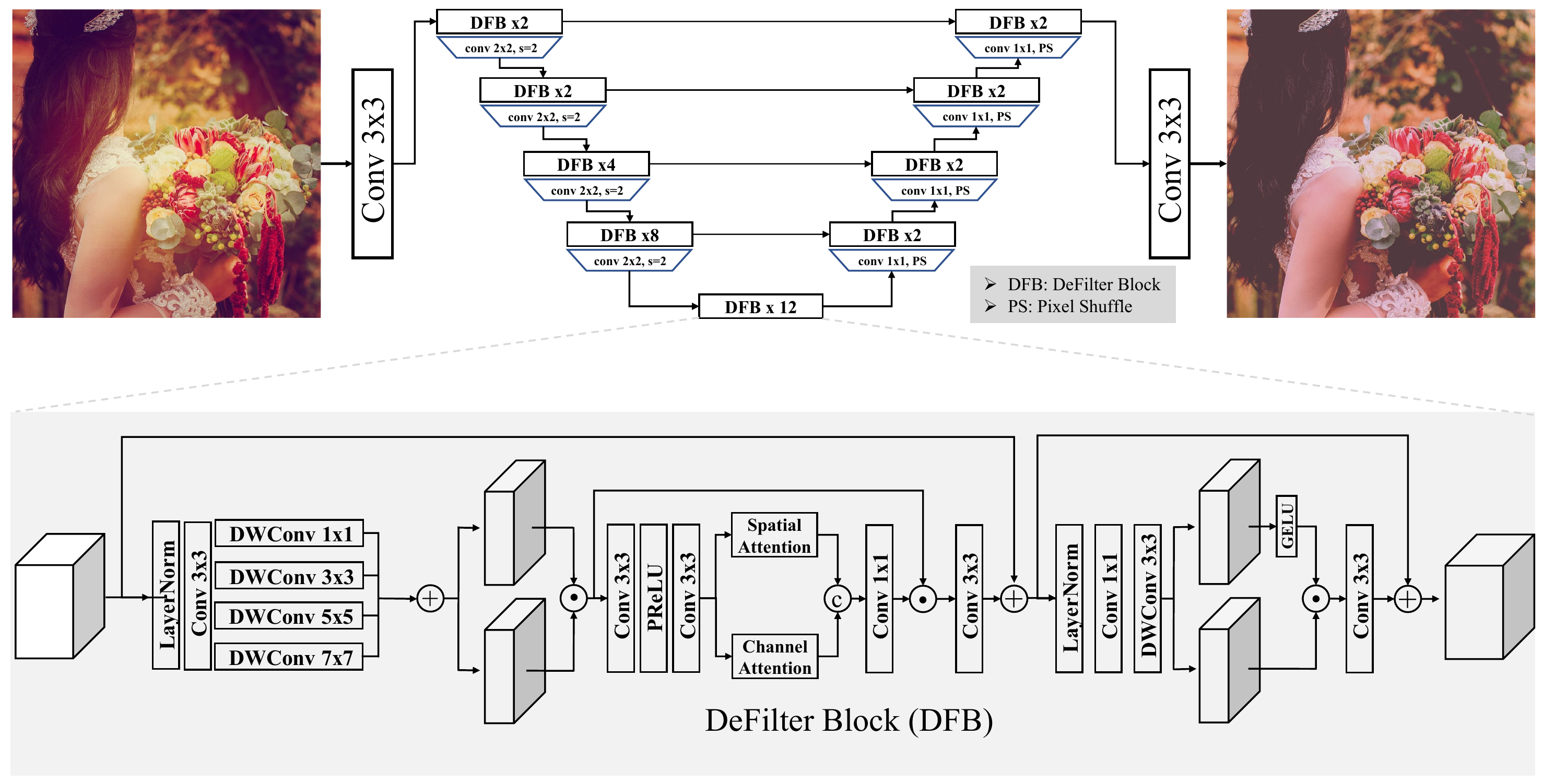}
      \caption{ The architecture of DefilterNet proposed by CASIA LCVG.}  
      \label{DefilterNet} 
      \end{center}
\end{figure}

\subsubsection{MiAlgo}
The MiAlgo team proposed a Two-stage Network for the Instagram Filter Removal task. In the first stage, we adopt the BigUNet to process the low-resolution input images. The BigUNet is improved based on MWCNN and RCAN. We replaced the convolutional layer in MWCNN with the residual group (RG, with channel attention layer) in RCAN to enhance the reconstruction ability of the network. In the second stage, we downsample the high-resolution input, then concatenate it with the output of the first stage, and feed it into the refine network. The refine network contains 5 residual groups with channel attention and spatial attention, which can get better details, as well as more accurate local color, local brightness, etc.

The BigUNet is first trained for 4000 epochs with a learning rate of $1e-4$. Then freeze the parameters of BigUNet, and train the refine network for 2000 epochs with a learning rate of $1e-4$. Finally, the overall finetune is trained for 2000 epochs with a learning rate of $1e-5$. All stages use Charbonnier loss and use cosine annealing for learning rate decay, the batch size is set to 16, and the whole image is used for training. In the early stage of training, the filter style is randomly selected; in the later stage of training, we increase the weight of the hard styles, that is hard example mining. At test time, we fuse the outputs of the first and second stages.

\begin{figure}[t]
   \begin{center}
     \includegraphics[width=0.6\textwidth]{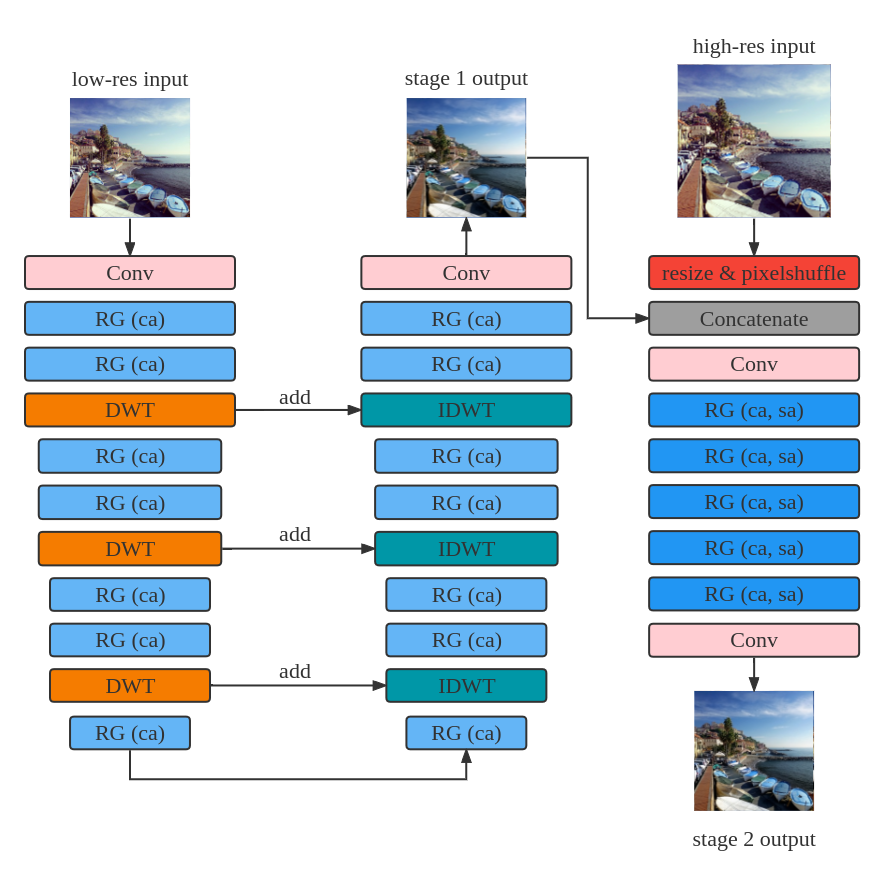}
      \caption{ The architecture of Two-stage Network proposed by MiAlgo.}  
      \label{DefilterNet} 
      \end{center}
\end{figure}

\subsubsection{Strawberry}
The solution of Strawberry is composed of different architectures, such as Uformer \cite{wang2022uformer}, Pix2Pix \cite{pix2pix2017}, VRT \cite{liang2022vrt}, RNN-MBP-Local \cite{chao2022}, and CycleGAN \cite{CycleGAN2017}. The pre-training weights are not used for the experiments. In the early stage of training, the team simply enhanced the data, processed brightness and contrast, and introduced random noise. No additional data is used in the experiments. The method is the fusion of the result after training several main models, and different models need to be trained separately. The test-time augmentation method is preferred for a single model to dynamically fuse the results. 

\subsubsection{SYU-HnVLab}
\begin{figure*}[t]
    \centering
    \begin{subfigure}[b]{0.45\textwidth}
        \centering
        \includegraphics[width=\textwidth]{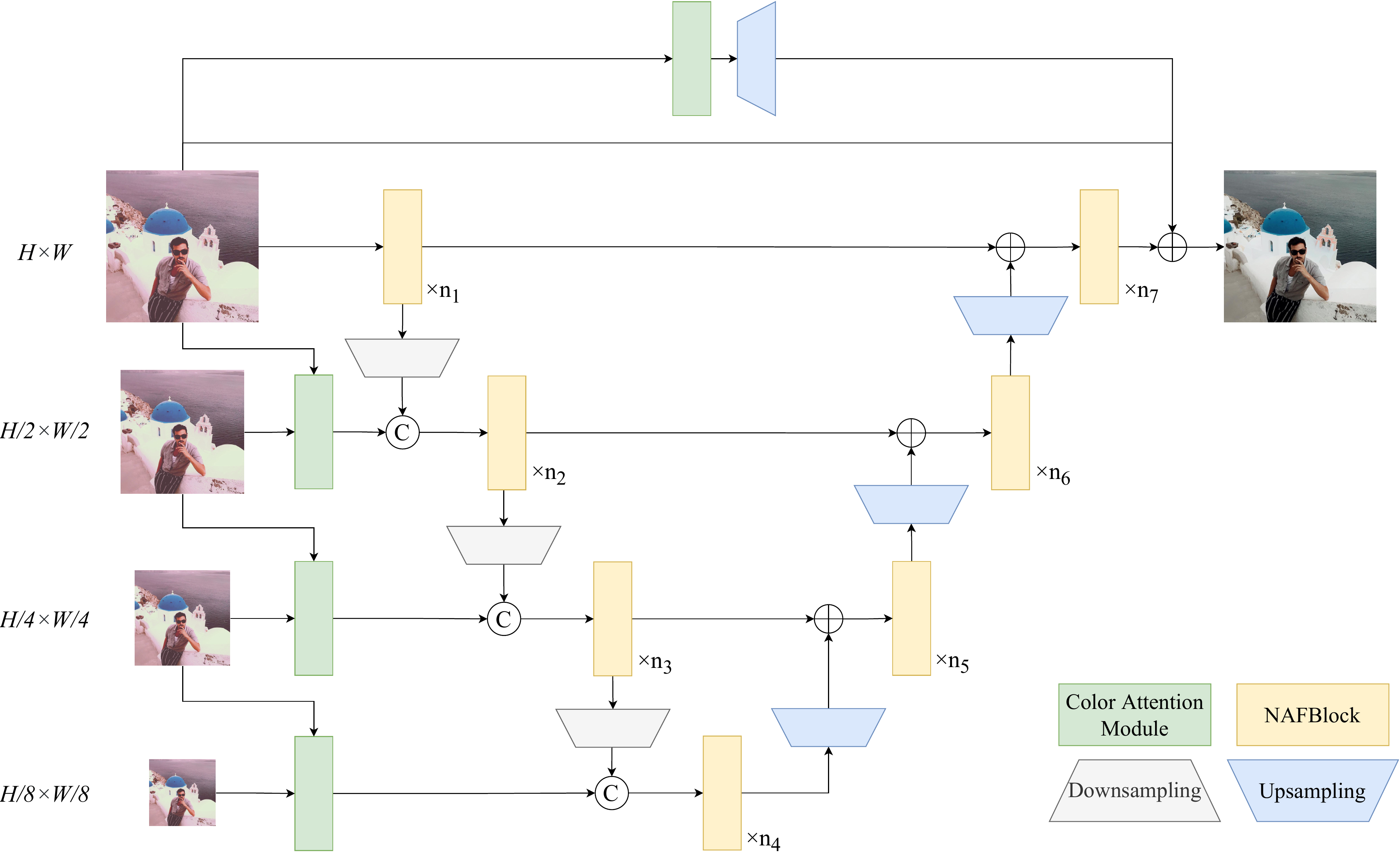}
        \caption{(a)}
    \end{subfigure}
    \begin{subfigure}[b]{0.45\textwidth}
        \centering
        \includegraphics[width=\textwidth]{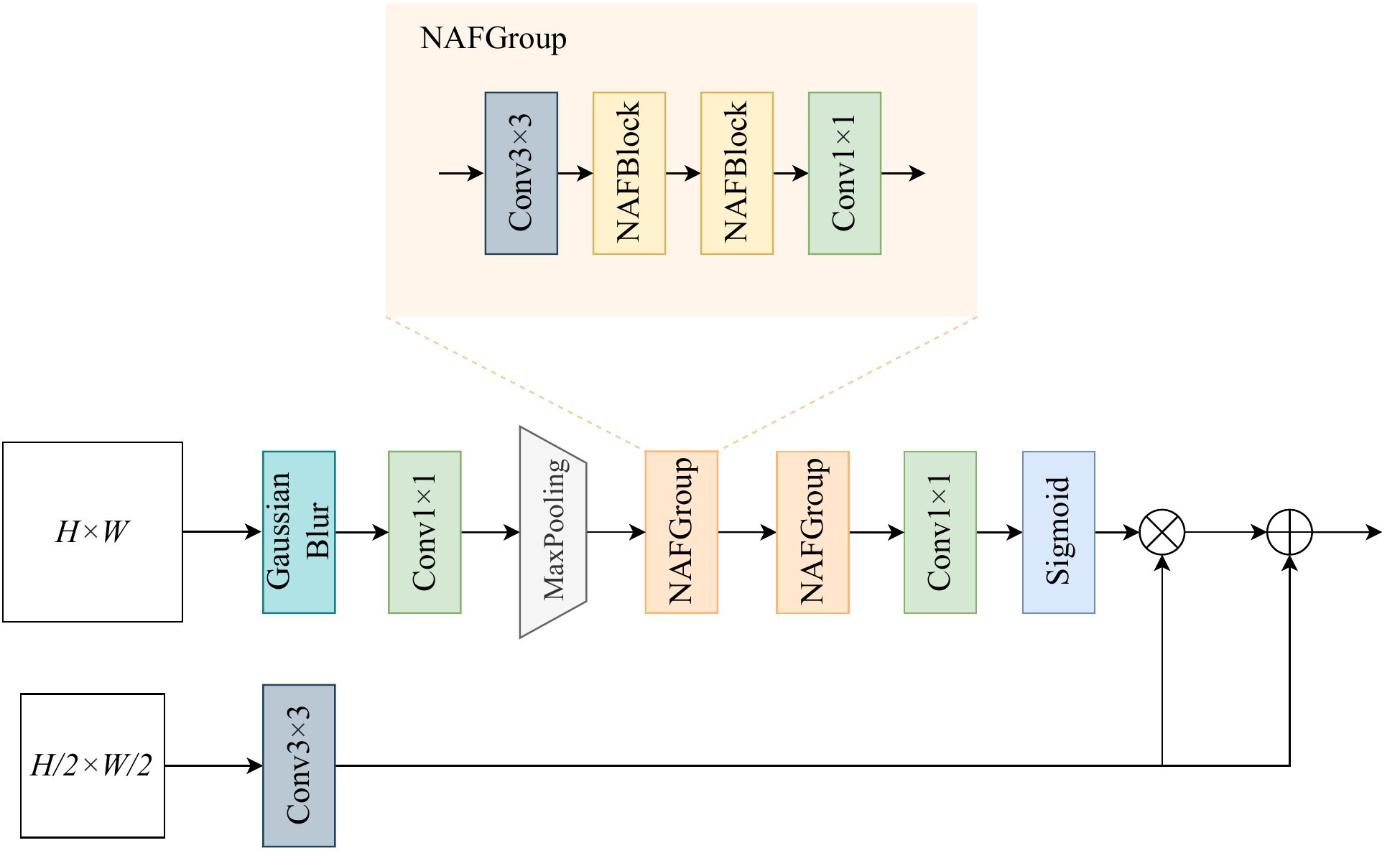}
        \caption{(b)}
    \end{subfigure}
    \par\bigskip
    \begin{subfigure}[b]{0.6\textwidth}
        \centering
        \includegraphics[width=\textwidth]{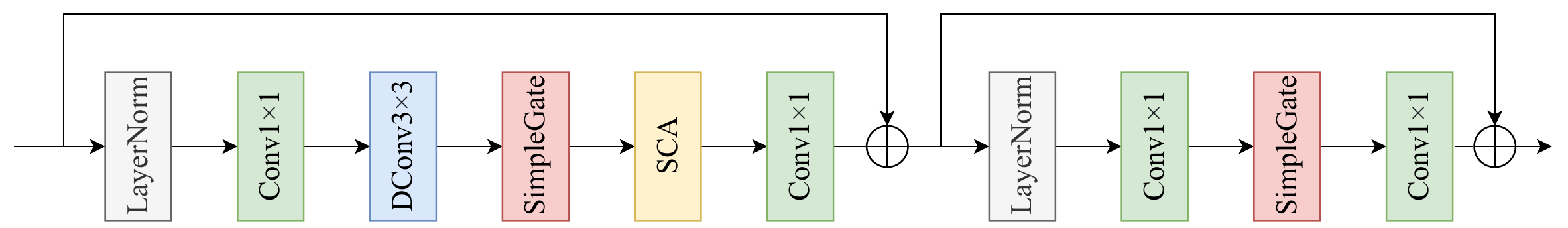}
        \caption{(c)}
    \end{subfigure}
    \caption{The proposed solution by SYU-HnVLab Team. \textbf{(a)} The main architecture of the proposed multi-scale color attention network (CAIR). \textbf{(b)} The proposed color attention module. {$\bigotimes$: element-wise multiplication, $\bigoplus$: element-wise summation. \textbf{(c)} NAFBlock~\cite{chen2022simple}}}
    \label{fig:syu-hnvlab}
\end{figure*}

The SYU-HnVLab team proposed Multi-Scale Color Attention Network for Instagram Filter Removal (CAIR) \cite{yeo2022cair}. The proposed CAIR is modified from NAFNet \cite{chen2022simple} with two improvements: 1) Color attention (CA) mechanism inspired by the color correction scheme of CycleISP \cite{zamir2020cycleisp}, and 2) CAIR takes multi-scale input images. The main difference between the filter image and the original image is the \textit{color} of images. Therefore, it is essential to consider color information to remove filters from filter images effectively. Through the CA mechanism, useful color information can be obtained from the multi-scale input images. We downscale an input image ($H\times W$) by a factor of 2 in each level, as described in Figure \ref{fig:syu-hnvlab}. The input image of each level and the higher level image pass together through the color attention module, and the color attentive features are obtained. These features are concatenated with the features from higher-level and then passes through NAFBlock. NAFBlock and the modules in NAFBlock are described in Figure \ref{fig:syu-hnvlab}. The number of NAFBlocks of each level ($n_1$-$n_7$) is set the same as NAFNet's setting, which is 2, 2, 4, 22, 2, 2, 2, respectively. The color attention module consists of Gaussian blur with $\sigma=12$ module, two $1\times1$ convolution layers, and two NAFGroup, comprised of two convolution layers and two NAFBlocks. The proposed network shows better results with lower computational and memory complexity when using the color attention mechanism than the original NAFNet under the same configuration. 

Before training the model, we resize high-resolution (HR) images ($1080\times 1080$) into $256\times 256$ due to a lack of training dataset. For resizing, an image is interpolated through resampling using pixel area relation (\textit{i}.\textit{e}., OpenCV interpolation option INTER\_AREA) for image quality.
As a result, we use three sets of images including HR images cropped in 256, resized HR images, and the low resolution ($256\times 256$) images. This data augmentation helps improve the flexibility of the trained model. 
In addition, images are randomly flipped and rotated with the probability of 0.5 in the training process. We use AdamW \cite{loshchilov2017decoupled} optimizer with $\beta_1=0.9, \beta_2=0.9$, weight decay $1e^{-4}$. The initial learning rate is set to $1e^{-3}$ and gradually reduces to $1e^{-6}$ with the cosine annealing schedule~\cite{loshchilov2016sgdr}. It is trained for up to 200K iterations. The training mini-batch size is 64, and the patch size is $256\times256$. We use Peak Signal-to-Noise Ratio (PSNR) metric as the loss function, which is PSNR loss. The number of channels in each NAFBlock is set to 32. As in~\cite{timofte2016seven,lim2017enhanced}, we use the self-ensemble method for test time augmentation (TTA), which produces seven new variants for each input. Therefore, each filter removal image is obtained by taking the average of all outputs processed by the network from augmented inputs. Additionally, we improve the output of CAIR through ensemble learning.


\begin{figure*}[t]
  \centering
  \begin{subfigure}[b]{0.8\textwidth}
        \centering
        \includegraphics[width=\textwidth]{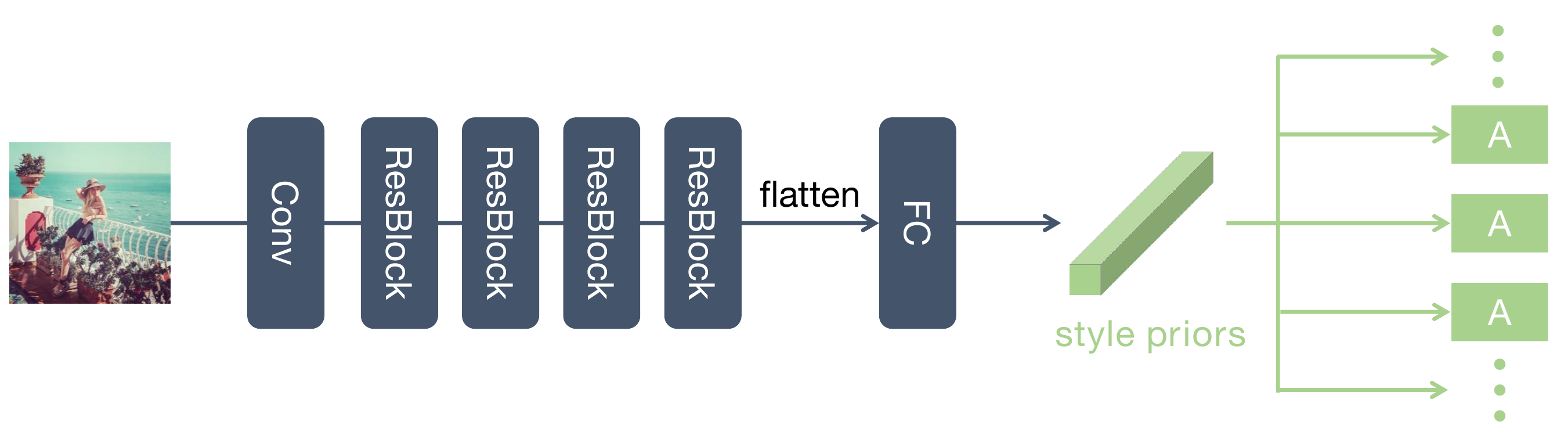}
        \caption{(a)}
        \label{xder-style}
    \end{subfigure}
    \begin{subfigure}[b]{0.49\textwidth}
        \centering
       \includegraphics[width=\textwidth]{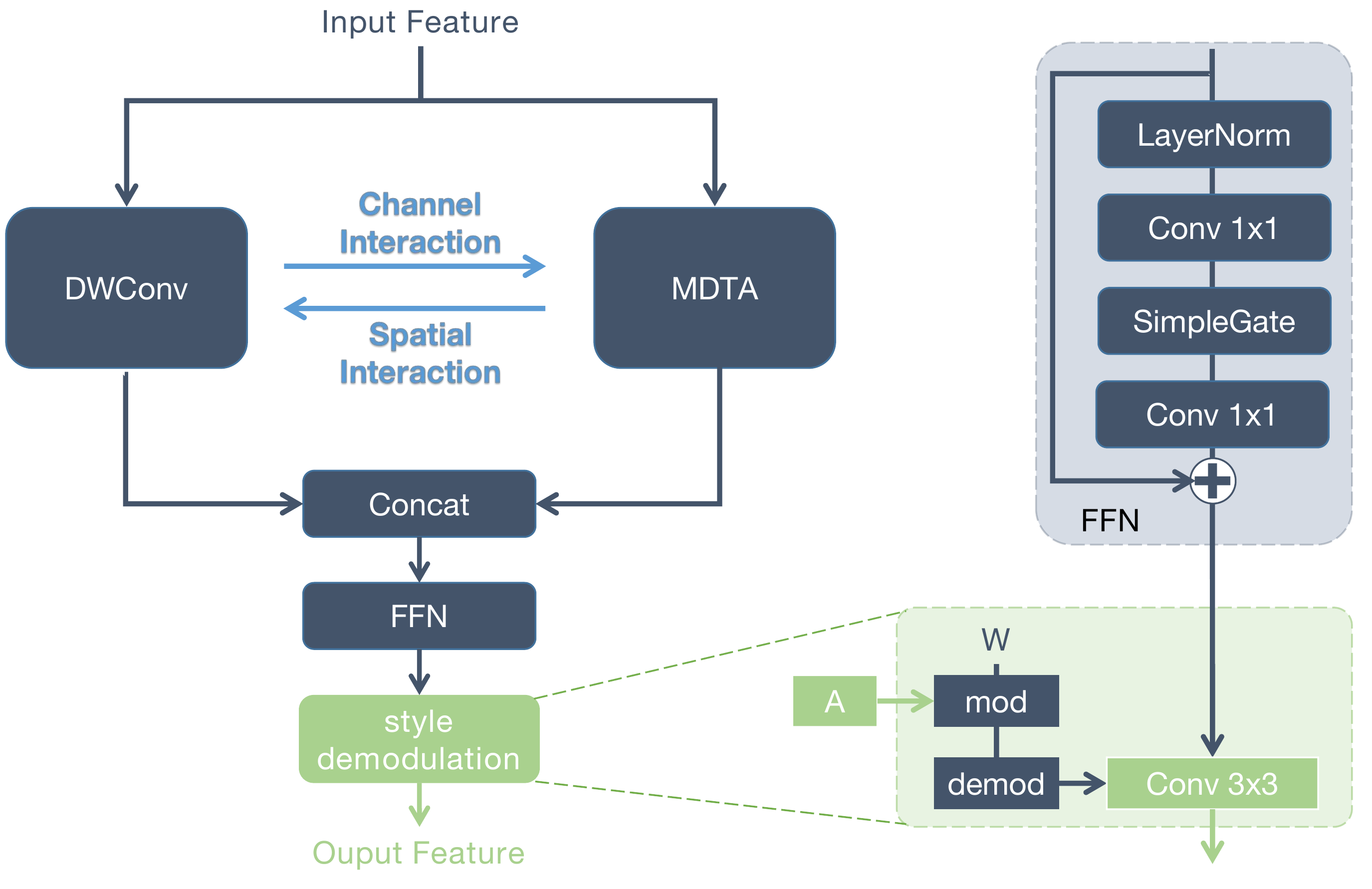}
        \caption{(b)}
        \label{xder-styleformer-block}
    \end{subfigure}
    \begin{subfigure}[b]{0.49\textwidth}
        \centering
       \includegraphics[width=\textwidth]{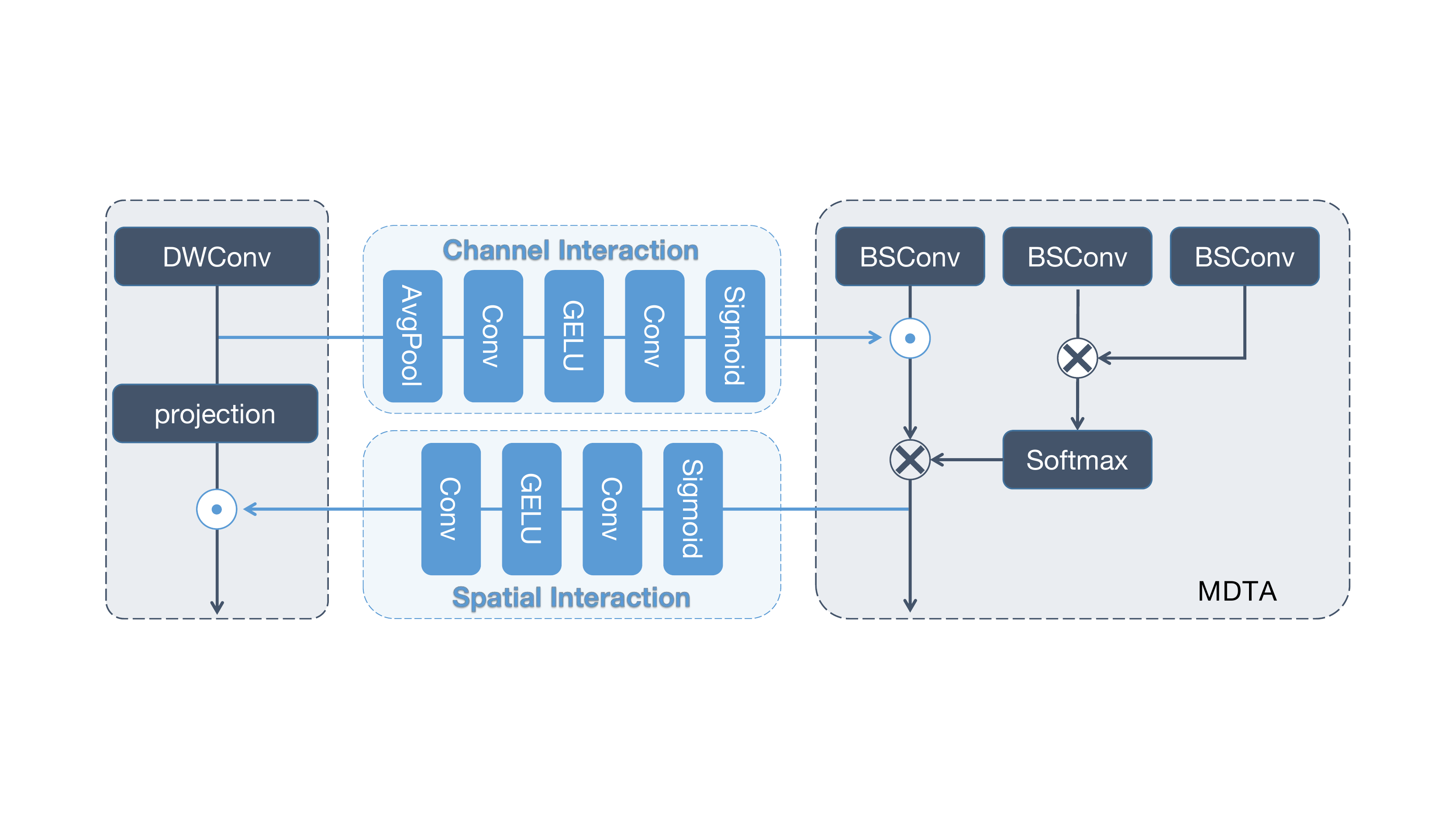}
        \caption{(c)}
        \label{xder-interaction}
    \end{subfigure}
    \caption{The proposed solution by XDER team. \textbf{(a)} The style priors extraction network. \textbf{(b)} The interaction mechanism. \textbf{(c)} The StyleMixformer Block. $\bigodot$: element-wise multiplication, $\bigotimes$: matrix multiplication, $\bigoplus$: element-wise addition. $A$ denotes a learned affine transform from style priors.}
\end{figure*}

\subsubsection{XDER}
The XDER team proposed a StyleMixFormer network for the Instagram Filter Removal task. Inspired by Kınlı \textit{et al.} \cite{Kinli_2021_CVPR}, we consider this problem as a multi-style conversion problem by Assuming that each filter introduces global and local style manipulations to the original image. In order to extract style priors information from the filtered image, we design a style priors extraction module as shown in Figure \ref{xder-style}. Compared with AdaIN, we introduce a more effective style demodulation \cite{Karras2019stylegan2} to remove style of filtered image. As shown in Figure \ref{xder-styleformer-block}, we propose a style demodulation Mixing block in order to fully extract the features of the filtered image and transform the style of the feature map. Inspired by Chen \textit{et al.} \cite{chen2022mixformer}, we extract global and local features by attention mechanism and depth-wise convolution, respectively, and interact information between the two features. The details of the information interaction are shown in Figure \ref{xder-interaction}. Unlike Chen \textit{et al.} \cite{chen2022mixformer} we replace the local-window self-attention block with multi-Dconv head transposed attention (MDTA) \cite{Zamir2021Restormer} which has been experimentally shown to be effective on low-level vision. For the feed-forward network, we adopt the SimpleGate mechanism \cite{chen2022simple}. We use the Unet network as our overall architecture following Chen et al. \cite{chen2022simple}.

We training StyleMixFormer in a two stage manner. In the first stage, we use a 4-layer ResNet to extract the style priors of different filtered images with ArcFace loss \cite{deng2019arcface}, which can effectively narrow intra-class distances and increase inter-class distances. In the second stage, we train StyleMixFormer with L1 loss. For the second stage, we used the L1 loss and Adam optimizer for training. The initialized learning rate is 5e-4, the batch size is 16, and the total number of iterations is 600k. More detail and the source code can be found at \href{https://github.com/clearlon/StyleMixFormer}{https://github.com/clearlon/StyleMixFormer}.

\subsubsection{CVRG}

\begin{figure*}[t]
\centering
\includegraphics[width=\linewidth,keepaspectratio]{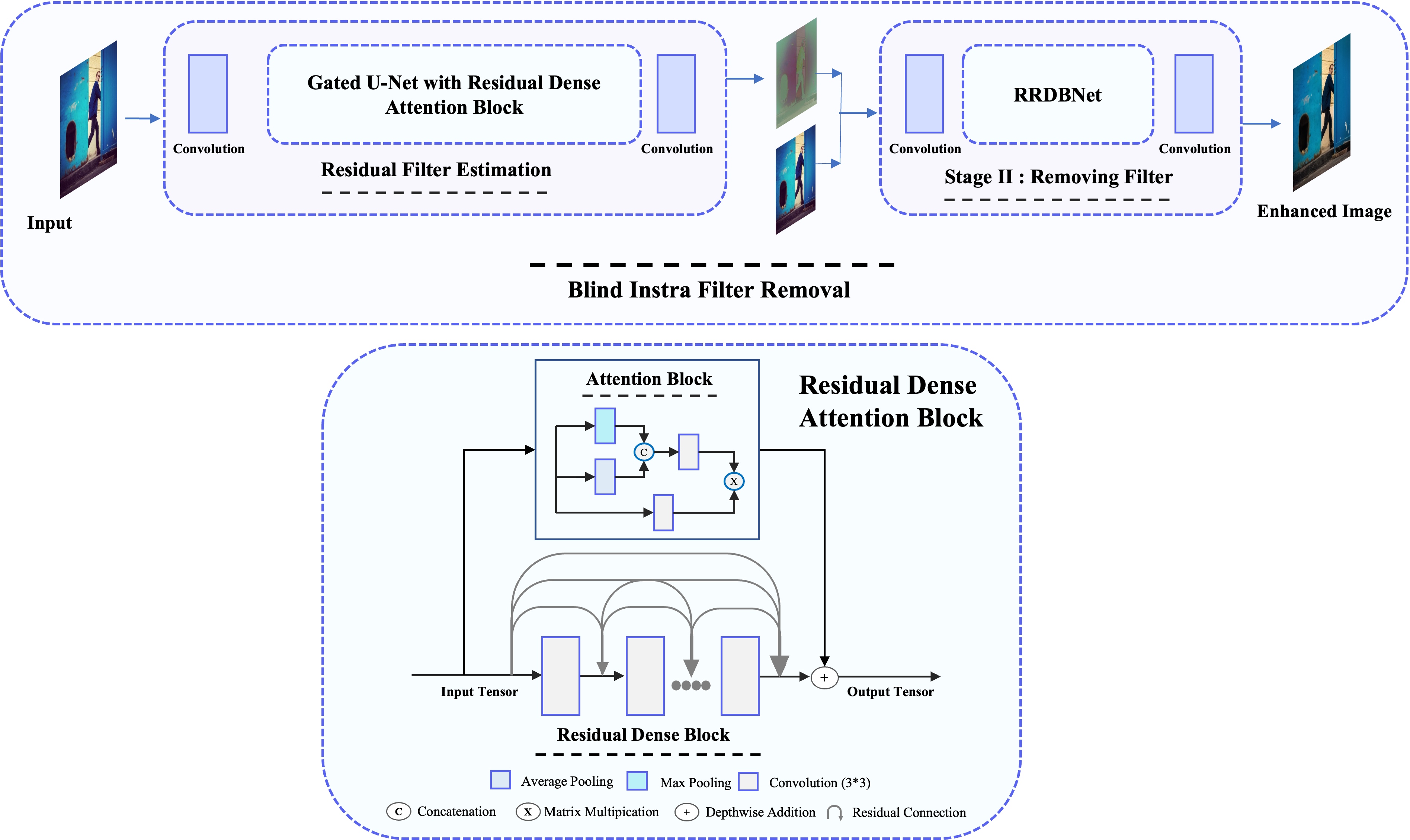}
\caption{ The overview of two-stage blind Instagram filter removal proposed by CVRG.  }
\label{fig:cvrg_overview}
\end{figure*}

Figure \ref{fig:cvrg_overview} illustrates the overview of the proposed two-stage solution. CVRG Team has developed the stage-I based on Gated U-Net \cite{sharif2021sagan,sharif2022deep} with residual dense attention block \cite{a2021two}. 
Stage-II of the solution comprises residual-in-residual dense blocks (RRDB) \cite{wang2018esrgan}. It leverages the residual filters (from stage-I) and the input images to obtain the final output. The $\mathcal{L}_1$ + Gradient loss was utilized to optimize both stages. Also, both networks were optimized with  Adam optimizer \cite{kingma2014adam}, where the hyperparameters were tuned as $\beta_1 = 0.9$, $\beta_2 = 0.99$, and learning rate $= 5e-4$.  We trained our model for 65 epochs with a constant batch size of 4. It takes around four days to complete the training process. We conducted our experiments on a machine comprised of an AMD Ryzen 3200G central processing unit (CPU) clocked at 3.6 GHz, a random-access memory of 16 GB, and n single Nvidia Geforce GTX 1080 (8GB) graphical processing unit (GPU).

\subsubsection{CVML}
The CVML team proposed a two-branch model for the Instagram Filter Removal task. 
One network learns an intrinsic filter from an input image with a color palette.
Another network learns unfiltered images directly from the input images. 
This approach fuses each unfiltered image to get the final result.
The proposed model has the advantage of being lightweight, fast inference time, and fine quality for resulting images.

\begin{figure*}[t]
    \centering
    \includegraphics[width=0.9\linewidth]{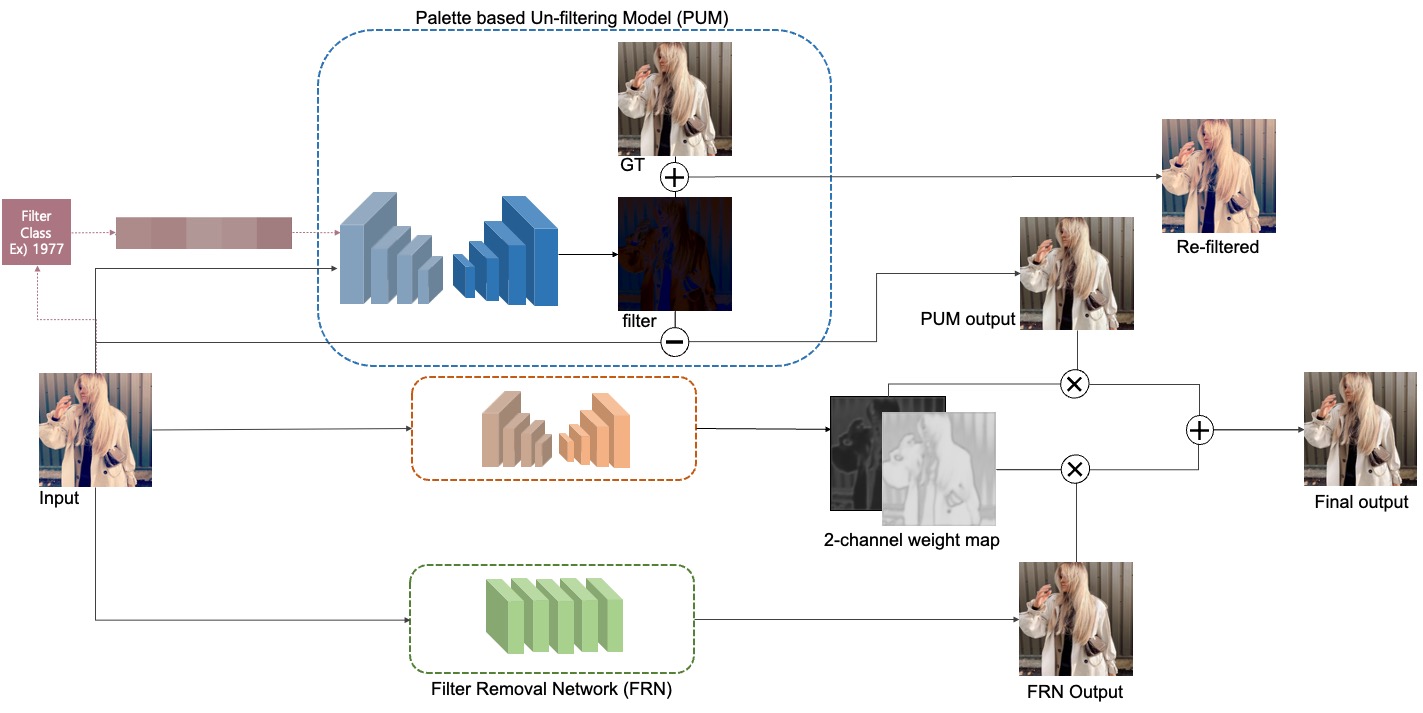}
  \caption{\textbf{Overall architecture of the CVML. \textit{i.e.}, two-branch model}. 
  The model removes the filter from the input image in two ways and integrates each result. 
  }
 \label{fig:arch}
\end{figure*}

The model named Palette-based Un-filtering Model (PUM) is a convolutional model with an encoder-decoder structure. 
The palette and the filtered images are inserted and the model learns an intrinsic Instagram filter. 
The palette has colors representing the corresponding filter.
The intrinsic filter is obtained as the output of the network and subtracted from the filtered image to obtain an unfiltered PUM output image.
The intrinsic filter is used once again to recreate the filtered image by adding to the ground truth image.
With a re-filtered image, the missing features can be checked in the intrinsic filter.
The color palette is generated by extracting five average colors with the K-means algorithm.
In this stage, all corresponding images of each filter are added.
The Filter Removal Network (FRN), is a model of a simple encoder-decoder structure, which creates an unfiltered image directly from a filtered image.
Output images from PUM and FRN are fused with a 2-channel weight map to obtain the final result.
The weight map is used to determine which portion of each output will be used for the final output. 

This model is trained for 300,000 steps.
The initial learning rate is set to $1e-4$ and is reduced by the cosine annealing strategy.
It used IFFI (Instagram Filter Removal on Fashionable Images) dataset images with low-resolution (256x256) for the experiments and a color palette generated for each filter.
This solution has performed data augmentation through 90, 180, and 270 degrees rotation.
It takes about 1 day to train our model and about 0.025 sec. per image to test.

\subsubsection{CougerAI}  
\begin{figure}[t]
  \centering
     \includegraphics[width=1\textwidth]{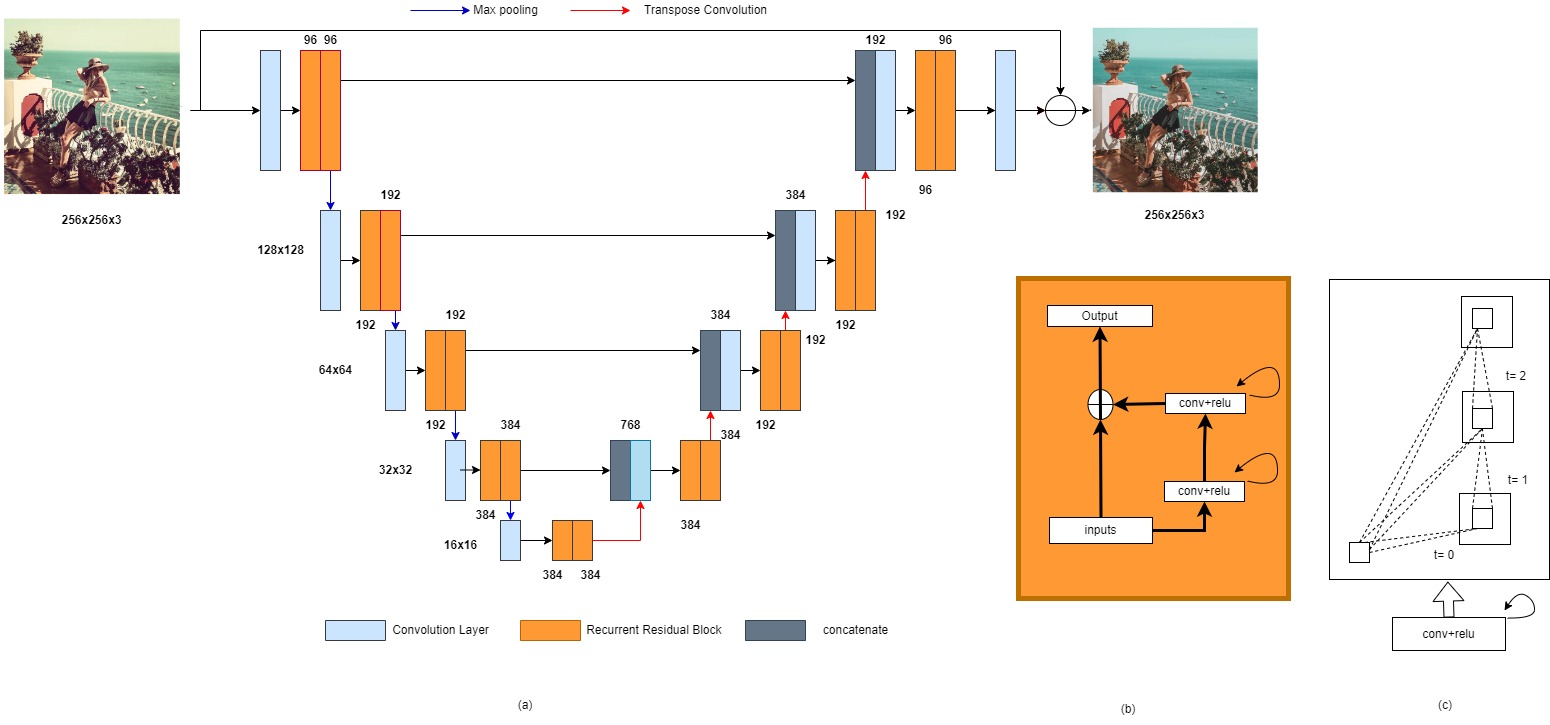}
  \caption{(a) Overall architecture of the Insta Net; (b) Typical Recurrent Residual Unit (c) Unfolding of a RCL Unit for timestep = 2.}
  \label{fig:CougerAI}
\end{figure}

As shown in Figure ~\ref{fig:CougerAI}, the proposed architecture is a U-Net-based structure that includes an encoding path to down-sample features and a decoding path to retrieve the required image. In each step of the encoding path, the input features are passed to a convolutional layer with $3\times3$ filter and then two recurrent residual blocks \cite{alom2018recurrent,liang2015recurrent} with time step 2. The output of the recurrent residual block is down-sampled by the factor of two using max pooling operation. The image is down-sampled 5 times in the encoding path. The same structure is used in the decoding path, where the up-sampled output is concatenated with the same-step output of the encoding path using a skip connection. After that, it is passed to the convolutional block and two subsequent recurrent residue blocks like the encoding path. As we have taken this task as an image restoration task, we subtracted the input image to retrieve the original unfiltered image.
For the purpose of training, the images are normalized between 0 to 1. To update the weights during training, we used the Adam optimizer \cite{DBLP:journals/corr/KingmaB14}. The learning rate is initialized with 0.001 and reduced by 10 percent after 15 epochs if the validation loss does not improve. The batch size is set to 2. The model is evaluated using the Peak-Signal-to Noise Ratio (PSNR) and Structural Similarity Index (SSIM). The model is trained for 200 epochs on a single 16 GB NVIDIA Tesla K80 GPU in Google Colab pro.

\section{Teams and Affiliations}

\subsection{Organizers of AIM 2022 Challenge on Instagram Filter Removal}

\begin{itemize}
    \item  Furkan Kınlı\textsuperscript{1}, Sami Menteş\textsuperscript{1}, Barış Özcan\textsuperscript{1}, Furkan Kıraç\textsuperscript{1}, Radu Timofte\textsuperscript{2}
\end{itemize}

\textit{Affiliations:}
\begin{itemize}
    \item \textsuperscript{1} Özyeğin University, Türkiye.
    \item \textsuperscript{2} University of Würzburg, Germany.
\end{itemize}

\subsection{Fivewin}

\begin{itemize}
  \item Yi Zuo, Zitao Wang, Xiaowen Zhang
\end{itemize}
\textit{Affiliations:}
\begin{itemize}
    \item IPIU Laboratory, Xidian University.
\end{itemize}

\subsection{CASIA LCVG}

\begin{itemize}
  \item Yu Zhu\textsuperscript{1}, Chenghua Li\textsuperscript{1}, Cong Leng\textsuperscript{1,2,3}, Jian Cheng\textsuperscript{1,2,3}
\end{itemize}
\textit{Affiliations:}
\begin{itemize}
    \item \textsuperscript{1}Institute of Automation, Chinese Academy of Sciences, Beijing, China.
    \item \textsuperscript{2}MAICRO, Nanjing, China.
    \item \textsuperscript{3}AiRiA, Nanjing, China.
\end{itemize}

\subsection{MiAlgo}

\begin{itemize}
  \item Shuai Liu, Chaoyu Feng, Furui Bai, Xiaotao Wang, Lei Lei
\end{itemize}
\textit{Affiliations:}
\begin{itemize}
    \item Xiaomi Inc., China.
\end{itemize}

\subsection{Strawberry}

\begin{itemize}
  \item Tianzhi Ma, Zihan Gao, Wenxin He
\end{itemize}
\textit{Affiliations:}
\begin{itemize}
    \item Xidian University.
\end{itemize}

\subsection{SYU-HnVLab}

\begin{itemize}
  \item Woon-Ha Yeo, Wang-Taek Oh, Young-Il Kim, Han-Cheol Ryu
\end{itemize}
\textit{Affiliation:}
\begin{itemize}
    \item Sahmyook University, Seoul, South Korea.
\end{itemize}

\subsection{XDER}

\begin{itemize}
  \item Gang He, Shaoyi Long
\end{itemize}
\textit{Affiliations:}
\begin{itemize}
    \item Xidian University, Xi’an Shaanxi, China.
\end{itemize}

\subsection{CVRG}

\begin{itemize}
  \item S. M. A. Sharif, Rizwan Ali Naqvi, Sungjun Kim
\end{itemize}
\textit{Affiliations:}
\begin{itemize}
    \item FS Solution, South Korea.
    \item Sejong University, South Korea.
\end{itemize}

\subsection{CVML}

\begin{itemize}
  \item Guisik Kim, Seohyeon Lee
\end{itemize}
\textit{Affiliations:}
\begin{itemize}
    \item Chung-Ang University, Republic of Korea.
\end{itemize}

\subsection{Couger AI}

\begin{itemize}
  \item  Sabari Nathan,  Priya Kansal
\end{itemize}
\textit{Affiliations:}
\begin{itemize}
    \item Couger Inc, Japan.
\end{itemize}

\section{Acknowledgements}

We thank the sponsors of the AIM and Mobile AI 2022 workshops and challenges: AI Witchlabs, MediaTek, Huawei, Reality Labs, OPPO, Synaptics, Raspberry Pi, ETH Z\"urich (Computer Vision Lab) and University of W\"urzburg (Computer Vision Lab).

%
%
\bibliographystyle{splncs04}
\bibliography{egbib}
\end{document}